\documentclass[runningheads]{llncs}
\usepackage[T1]{fontenc}
\usepackage{booktabs}      %
\usepackage[table]{xcolor} 
\usepackage{graphicx,verbatim}
\usepackage{xcolor}
\usepackage{amsmath} 
\usepackage{amssymb}
\usepackage{multirow}
\usepackage{hyperref}
\usepackage{caption}
\newcommand{\myparagraph}[1]{
\noindent
\textbf{#1} ---
}

\begin{document}
\title{From Patches to Patients: A study of the tile-to-slide performance transferability in Digital Pathology}
\titlerunning{From Patches to Patients: Tile-to-slide transferability in Digital Pathology}
\begin{comment}  %% Removed for anonymized MICCAI submission
\author{First Author\inst{1}\orcidID{0000-1111-2222-3333} \and
Second Author\inst{2,3}\orcidID{1111-2222-3333-4444} \and
Third Author\inst{3}\orcidID{2222--3333-4444-5555}}
%
\authorrunning{F. Author et al.}
\institute{Princeton University, Princeton NJ 08544, USA \and
Springer Heidelberg, Tiergartenstr. 17, 69121 Heidelberg, Germany
\email{lncs@springer.com}\\
\url{http://www.springer.com/gp/computer-science/lncs} \and
ABC Institute, Rupert-Karls-University Heidelberg, Heidelberg, Germany\\
\email{\{abc,lncs\}@uni-heidelberg.de}}

\end{comment}

\author{Sofiène Boutaj\inst{1,2} \and
Leo Fillioux\inst{1,2} \and
Maria Vakalopoulou\inst{1,2} \and
Stergios Christodoulidis\inst{1,2}$^{\dagger}$\ \and
Pierre Marza\inst{1,2}$^{\dagger}$}

%index{Boutaj, Sofiène}
%index{Fillioux, Leo}
%index{Vakalopoulou, Maria}
%index{Christodoulidis, Stergios}
%index{Marza, Pierre}

\authorrunning{S. Boutaj et al.}

\institute{Université Paris-Saclay, CentraleSupélec, Gustave Roussy, INSERM, IHU PRISM, Cancer Data Science Unit, France \and Université Paris-Saclay, CentraleSupélec, MICS Laboratory, France \\
Corresponding author: \email{sofiene.boutaj@centralesupelec.fr} }

%index{Boutaj, Sofiène}
%index{Fillioux, Leo}
%index{Vakalopoulou, Maria}
%index{Christodoulidis, Stergios}
%index{Marza, Pierre}
  
\maketitle              % typeset the header of the contribution

\def\thefootnote{$\dagger$}\footnotetext{Denotes equal contribution.}
\begin{abstract}

Foundation Models (FMs) have recently redefined the state-of-the-art in histopathology by providing robust representations for whole-slide image (WSI) analysis. However, selecting the optimal foundation model (FM) for a specific clinical cohort currently requires multiple pre-processing steps, followed by computationally expensive feature extraction and the training of a Multiple Instance Learning (MIL) aggregator for every model.
In this work, we investigate whether efficient tile-level linear probing can serve as a reliable proxy for slide-level performance, reducing the need to run full slide-level pipelines for every candidate encoder. We benchmark 19 state-of-the-art FMs on 42 slide-level and 16 tile-level tasks, comparing tile probing metrics against slide-level outcomes using ABMIL and Mean Pooling aggregations. We observe a high correlation between tile and slide performance across varying task difficulties, indicating that encoder representation quality is the primary determinant of WSI success. Sensitivity analyses show that transferability is stable across models and is more influenced by cohort sizes and numbers of tiles per slide than by average task difficulty. 

We also measure the agreement in best performing models between tile and slide-level tasks, showing tile benchmarks reliably shortlist strong candidates. Overall, our study indicates that tile-level benchmarking provides an efficient and practical first step for narrowing down candidate models, while slide-level evaluation remains essential for final validation on clinical tasks.

\keywords{Digital pathology \and Foundation models \and Benchmark.}
% Authors must provide keywords and are not allowed to remove this Keyword section.

\end{abstract}

\begin{figure}[t]
    \centering
    \includegraphics[width=\linewidth]{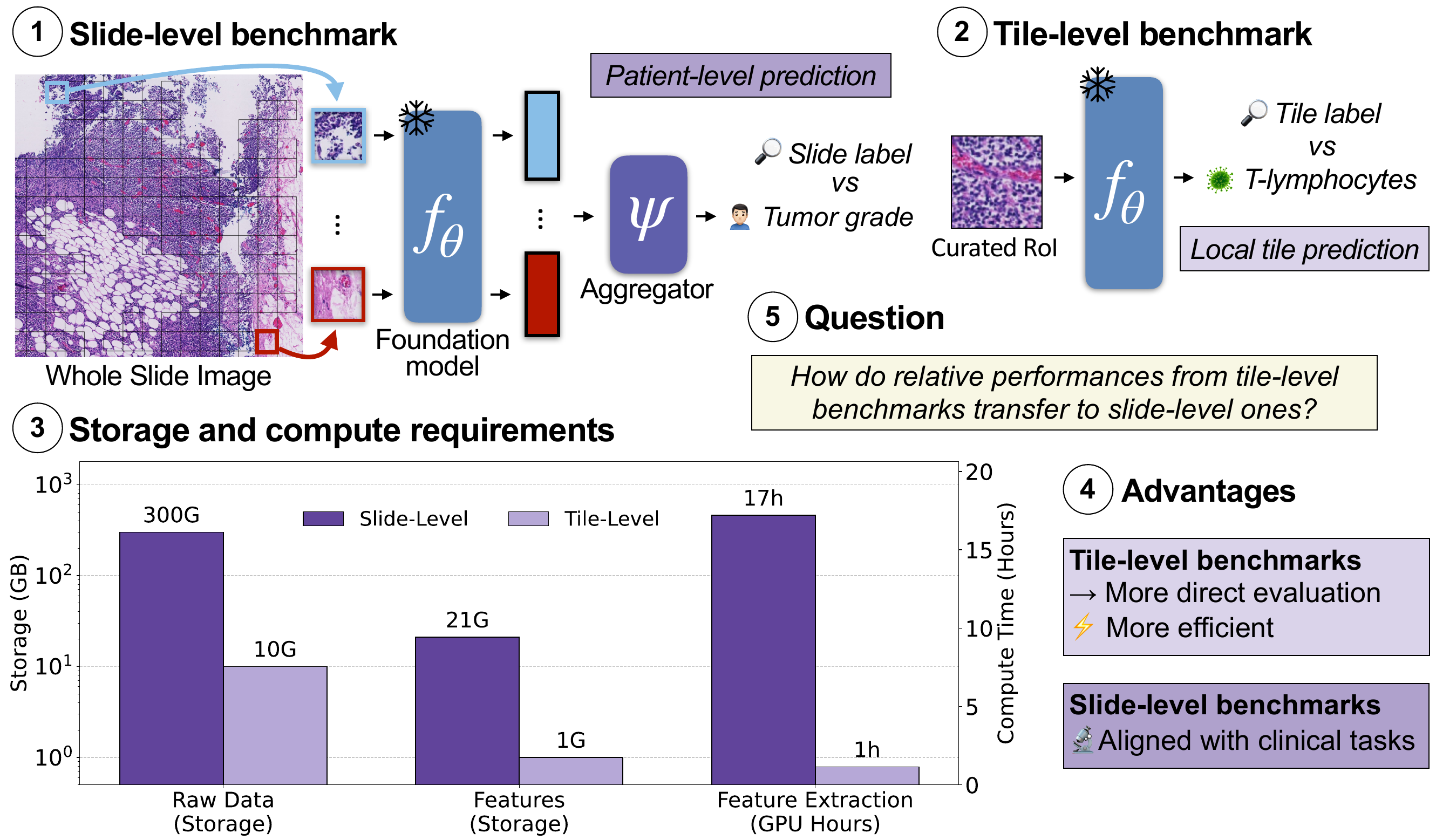}
    \caption{\textbf{Comparison of slide-level and tile-level benchmarks.} 
    \textbf{(1,2)} Overview of the $2$ benchmark types.
    \textbf{(3) Storage and compute requirements:} Average storage (log scale) per dataset and compute time per dataset and per model, measured on a single NVIDIA V100 GPU. Slide-level averages were computed across 42 tasks and 19 models, while tile-level averages were computed across 16 tasks and 19 models. } 
    \label{fig:teaser_fig}
\end{figure}

\section{Introduction}

Digital pathology plays a central role in diagnosis and prognosis by automatically extracting relevant information about the cellular environment from tissue imaging. This is particularly true now that many foundation models~\cite{chen2024towards,vorontsov2024foundation,zimmermann2024virchow2,hoptimus0,filiot2023scaling,filiot2024phikon} \cite{nechaev2024hibou,karasikov2025training,lu2024visual,ding2024multimodal,zhou2024knowledge,ikezogwo2023quilt,huang2023visual,xiang2025vision} were introduced as powerful generalist feature extractors for histopathology images. However, the high-resolution of Whole Slide Images (WSI) collected in histopathology makes it impossible to process them directly. There is a need for a specific processing pipeline: in practice, each WSI is first segmented to remove background and isolate tissue, then divided into tiles; each tile is embedded with a foundation model, and all tile-level features are aggregated to perform a slide-level prediction.

Comparing such foundation models, and better understanding their differences thus becomes important to draw a clear picture of the progress in the field but also from a practical point of view when selecting a feature extractor for a new clinical task. For these reasons, various benchmarks were proposed recently~\cite{wolflein2023benchmarking,kang2023benchmarking,neidlinger2024benchmarking,gustafsson2024evaluating,alfasly2024foundation,gatopoulos2024eva,breen2025comprehensive,lee2025benchmarking,majzoub2025good,alfasly2025validation,zhang2025accelerating,campanella2025clinical,marza2025thunder}. We can divide them in two categories: \textit{tile-level} and \textit{slide-level} benchmarks. The former evaluates how foundation model embeddings can extract relevant information from specific tiles, or regions of WSIs, while the latter targets slide-level predictions. Figure~\ref{fig:teaser_fig} provides an overview of both types of benchmarks, presents a comparison of their storage and compute efficiency, and summarizes their advantages. As presented in~\cite{marza2025thunder}, tile-level benchmarks are more efficient, isolate the impact of foundation models from required aggregators to perform slide-level predictions, and leverage denser supervision. On the other hand, slide-level benchmarks require more pre-processing, more compute, come with sparser supervision, and necessitate training aggregators on top of foundation model features, but are closer to the final clinical tasks of interest.

We are thus presented with the following dilemma: tile-level benchmarks allow a more direct and efficient evaluation of foundation models, which is appealing from a methodological point of view, while slide-level benchmarks better model clinical needs. An important question is thus: \textit{How do relative performances from tile-level benchmarks transfer to slide-level ones?} A good transfer would imply that tile-level benchmarks are effective tools for evaluating new models, providing confidence that their relative ranking will be preserved on slide-level tasks. However, this question is not binary but more nuanced. This paper aims not only to address it, but also to better understand the conditions under which performance transfers.

Our contributions are fourfold: (\textit{i}) to the best of our knowledge, we provide the first large-scale tile-to-slide benchmarking study across 19 open-source pathology foundation models, 16 tile-level tasks, and 42 slide-level tasks from publicly available datasets; (\textit{ii}) we measure tile-to-slide rank correlation using Pearson, Spearman, and Kendall metrics for both mean-pooling and ABMIL slide-level aggregation, and show strong transferability overall; (\textit{iii}) we perform sensitivity analyses showing stable correlations and highlighting the role of cohort size and number of tiles per slide for tile-to-slide transferability; and (\textit{iv}) we show through a top-5 overlap analysis that tile-level benchmarks are useful for shortlisting candidate encoders before expensive slide-level training, and highlight the current limitations of tile-level datasets.

\section{Related work}
\myparagraph{Foundation models in digital pathology} are trained in a self-supervised manner on large datasets spanning multiple organs, scanning protocols and centers. They can be divided into two categories: (\textit{i}) vision-only~\cite{chen2024towards,vorontsov2024foundation,zimmermann2024virchow2,hoptimus0,filiot2023scaling}, \cite{filiot2024phikon,nechaev2024hibou,karasikov2025training,vaidya2025molecular,wang2024pathology,xu2024whole} trained with DINO-style objectives~\cite{oquab2023dinov2}, and (\textit{ii}) vision-\hspace{0pt}language encoders~\cite{lu2024visual,ding2024multimodal,zhou2024knowledge,ikezogwo2023quilt,huang2023visual,xiang2025vision,shaikovski2024prism} optimized with CLIP-style losses~\cite{radford2021learning}. Such models are then used as powerful feature extractors for a variety of downstream tasks such as cancer sub-typing or gene mutation prediction.

\myparagraph{Benchmarking foundation models} With the recent surge of foundation models in digital pathology, systematic evaluation has become a bottleneck. This has led the community to search for methods of comparing them to answer two main questions: (\textit{i}) \textit{Which direction should the field take to build better-performing models?}, and (\textit{ii}) \textit{which models should practitioners use to get the best results?}
%The most common approach
A common approach is to benchmark foundation models on slide-level tasks~\cite{wolflein2023benchmarking,kang2023benchmarking,neidlinger2024benchmarking,gustafsson2024evaluating,alfasly2024foundation,gatopoulos2024eva,breen2025comprehensive,lee2025benchmarking,majzoub2025good,alfasly2025validation,zhang2025accelerating,campanella2025clinical}. These are clinically relevant, but require to couple a slide-level aggregator to existing tile-level foundation models to perform predictions. This induces additional computational constraints and also makes the assessment of the impact of the features from the foundation model itself less direct as many choices related to the design of the aggregator are to be made. Tile-level benchmarks~\cite{gatopoulos2024eva,marza2025thunder} are another alternative. They are generally more efficient and provide a more direct comparison of foundation model representation spaces, as they isolate their impact from any aggregation method. 

However, it remains unclear whether these two benchmarking paradigms lead to consistent conclusions when comparing foundation models. To the best of our knowledge, we present the first large-scale study of tile-to-slide (patch-to-patient) performance transferability.

\section{Patches to patients performance transferability}
\subsection{Experimental setup}

\myparagraph{Foundation Models}
We benchmark 19 recent pathology Foundation Models (FMs), covering both visual-only encoders and vision-language models. The set includes five vision-language models -- CONCH (v1, v1.5)~\cite{lu2024visual}, KEEP~\cite{zhou2024knowledge}, PLIP~\cite{huang2023visual}, QuiltNet~\cite{ikezogwo2023quilt} -- and 14 vision-only encoders -- ProvGigaPath~\cite{xu2024whole}, H-Optimus (0 and 1)~\cite{hoptimus0}, Hibou-Base~\cite{nechaev2024hibou}, H0-mini~\cite{filiot2025distilling}, Kaiko (ViT-S and ViT-B variants)~\cite{aben2024towards}, Phikon (v1, v2)~\cite{filiot2023scaling,filiot2024phikon}, UNI (v1, v2)~\cite{chen2024towards}, and Virchow (v1, v2)~\cite{vorontsov2024foundation}.

\myparagraph{Benchmarks and Tasks} For the tile-level evaluation, we include all patch-level \textit{classification} tasks from the \texttt{THUNDER} benchmark~\cite{marza2025thunder}, totaling 16 datasets (16 tasks) covering a broad range of tissue origins and tasks (e.g., tumor detection, histologic pattern recognition). In total, all datasets contain 2,202,752 tiles, with dataset sizes ranging from 408 to 367,229 samples. The details about the considered datasets are presented in~\cite{marza2025thunder} and on the \texttt{THUNDER} documentation\footnote{\href{https://mics-lab.github.io/thunder/\#available-datasets}{https://mics-lab.github.io/thunder/\#available-datasets}}. 

The slide-level analysis involves 42 tasks drawn from 19 WSI datasets (see Table~\ref{tab:slide_level_benchmark_summary}). Collectively, these slide-level datasets cover 10 anatomical sites (breast, lung, colon/rectum, kidney, ovary, uterus, brain, stomach, pancreas, and head \& neck). Following the standard \texttt{PathoBench} protocol~\cite{zhang2025accelerating}, we evaluate the \textit{CPTAC} tasks using a 50-fold train/test split, while employing 5-fold cross-validation for all other WSI datasets. For all splits, we ensure there is no data leakage between the training and test sets.

\begin{table*}[t]
\centering
\setlength{\tabcolsep}{4pt}
\caption{Detailed summary of the 19 slide-level WSI datasets (42 tasks), including average number of slides per task and average number of tiles per slide.}
\label{tab:slide_level_benchmark_summary}
\resizebox{\textwidth}{!}{
\begin{tabular}{cccccccc}
\toprule
\multirow{2}{*}{\textbf{Source}} &
\multirow{2}{*}{\textbf{Dataset}} &
\textbf{Mutation} &
\textbf{Molecular} &
\textbf{Histological} &
\textbf{Immune} &
\textbf{Avg Slides} &
\textbf{Avg Tiles} \\
& &
\textbf{tasks} &
\textbf{tasks} &
\textbf{tasks} &
\textbf{tasks} &
\textbf{/ Task} &
\textbf{/ Slide} \\
\midrule

CPTAC~\cite{edwards2015cptac} & 9 datasets
& 18 & 1 & 2 & 7
& 212
& 5089 \\

TCGA~\cite{tomczak2015review} & 8 datasets
& 8 & 2 & 1 & -
& 443
& 12919 \\

BRACS~\cite{brancati2022bracs} & 1 dataset
& - & - & 2 & -
& 545
& 13610 \\

CAMELYON16~\cite{ehteshami2017diagnostic} & 1 dataset
& - & - & 1 & -
& 398
& 14950 \\

\midrule

\textbf{Total} & \textbf{19 datasets}
& \textbf{26} & \textbf{3} & \textbf{6} & \textbf{7}
& \textbf{343}
& \textbf{9922} \\

\bottomrule
\end{tabular}}
\end{table*}

\myparagraph{Downstream tasks evaluation}
The tile-level tasks derived from \texttt{THUNDER}~\cite{marza2025thunder} are implemented as linear probing on frozen tile embeddings.
For slide-level tasks, we rely on the open-source \texttt{TRIDENT} library~\cite{zhang2025accelerating} to preprocess raw WSIs, including tissue segmentation, patching, and feature extraction. All WSIs are processed at $20\times$ magnification. Tile size is automatically selected according to the architectural requirements of each foundation model, using one of $\{224\times224,\;256\times256,\;512\times512\}$ pixels. Slide-level evaluation is performed with the \texttt{PathoBench} framework~\cite{zhang2025accelerating}, using its default implementations of two aggregation strategies: (\textit{i}) \textbf{mean pooling with linear probing}, where we compute a mean feature vector and train a linear classifier on the pooled vector; and (\textit{ii}) \textbf{attention-based MIL (ABMIL)}~\cite{ilse2018attention}, where we learn a gated attention mechanism that assigns a weight to each tile embedding and aggregates them into a slide representation, followed by a linear classification head (single attention head, projection dimension $512$, and pre-classification dropout $0.25$). We select ABMIL as it remains highly effective in practice, with prior studies demonstrating that its performance is often on par with more complex MIL architectures~\cite{shao2025multiple}, making it a robust and sufficient baseline for our evaluation. Overall, the pre-processing, feature extraction, and downstream training across all \textit{slide-level} tasks and foundation models required more than 15{,}000 V100 GPU-hours.

\myparagraph{Metrics}
For the slide-level evaluation, as it involves multiple datasets with varying numbers of tasks, we first compute the mean macro-F1 within each dataset, and subsequently average these intra-dataset means, preventing any single dataset from skewing the global WSI score, that we denote $S_m$. For the tile-level benchmark, the aggregated score $T_m$ is computed as the macro-F1 averaged across the 16 individual datasets in \texttt{THUNDER}.

\myparagraph{Conducted analyses}
Our main experiment is about (\textit{i}) quantifying the overall tile-to-slide transferability. For that, we analyze the relationship between the tile summary $T_m$ and the slide summary $S_m$ across all evaluated FMs. We report the correlation of performance values across models using three metrics: Pearson’s $\rho_P$ for the tile-to-slide performance linear relationship, and Spearman’s $\rho_S$ and Kendall’s $\tau$ for tile-to-slide rank-order consistency. We assess significance using a two-sided permutation test. Beyond this, we perform the following complementary analyses: (\textit{ii}) \textbf{Leave-one-model-out sensitivity analysis:} recompute the correlation after removing each FM to ensure results are not driven by a single model.
(\textit{iii}) \textbf{Task-ablation sensitivity:} track correlation as slide tasks are removed based on cohort size, number of tiles per slide, and average task performance to identify key dataset factors for tile-to-slide performance transferability.
(\textit{iv}) \textbf{Top-5 shortlist utility:} compute the overlap between top-5 tile-level and top-5 slide-level models for each tile/slide task pair.
(\textit{v}) \textbf{Rank-sum consensus:} aggregate ranks across Tile, Mean Pooling, and ABMIL to highlight consistently strong models and where reordering concentrates.
\subsection{Results}

\begin{figure}[t]
    \centering
    \includegraphics[width=\linewidth]{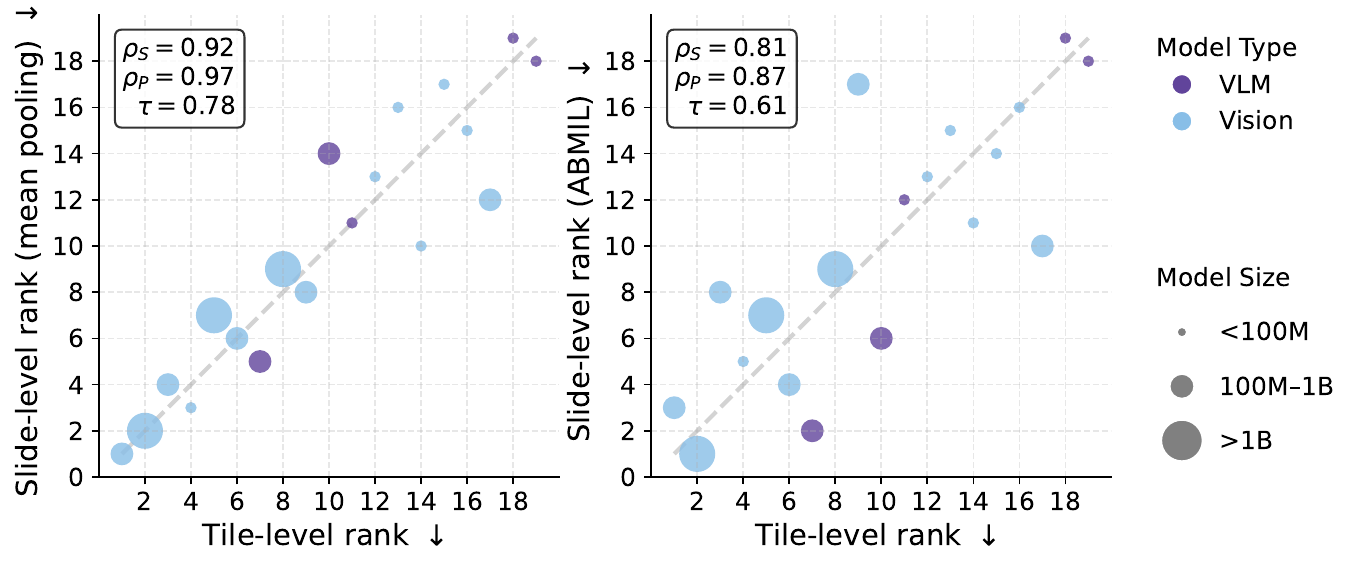}
    \caption{\textbf{Rank correlation between slide-level and tile-level benchmarks.} Comparing the rank across 19 models on a set of tile-level and slide-level tasks. Slide-level aggregation is performed via mean-pooling (\textit{left}) and ABMIL (\textit{right}); measured by Spearman ($\rho_S$), Pearson ($\rho_P$), and Kendall’s $\tau$.}
    \label{fig:main_figure}
\end{figure}

\myparagraph{Global rank correlation between tile-level and slide-level benchmarks} Fig.~\ref{fig:main_figure} reports the rank agreement between tile-level linear probing and slide-level performance across 19 foundation models. Using mean pooling for slide aggregation (\textit{left}), the model ordering is highly preserved, with points tightly concentrated around the identity line and strong correlations ($\rho_S=0.925$, $\tau=0.778$, $\rho_P=0.967$; permutation $p=2\times10^{-4}$ for all). This indicates that, under a simple aggregation scheme, slide-level success is largely determined by the intrinsic quality of the frozen tile representations. When switching to ABMIL (right), the correspondence remains significantly positive but is weaker ($\rho_S=0.814$, $\tau=0.614$, $\rho_P=0.874$; permutation $p=4\times10^{-4}$ for all), with larger departures from the diagonal. This suggests that learning a more expressive MIL aggregator introduces additional variability that can alter relative performance, particularly among mid-ranked models. The results are robust to the choice of evaluation metric: when replacing macro-F1 with balanced accuracy, the resulting correlations remain within a maximum absolute deviation of $\leq 2$\% from the macro-F1 correlations (with permutation p-values remaining highly significant). Moreover, the very high Pearson correlation ($\rho_P = 0.874$ for ABMIL and $\rho_P = 0.967$ for mean pooling) indicates a strong linear relationship between tile-level and slide-level performance. This confirms that tile-level benchmarking provides reliable quantitative information about downstream slide-level performance.  Overall, these results support tile-level probing as an efficient proxy for slide-level model benchmarking while highlighting that rank transferability depends on the complexity of the aggregation method. Importantly, as slide-level predictions are performed from aggregated tile embeddings, our results validate that high-quality tile representations (evaluated on tile-level benchmarks) naturally lead to strong slide-level performance.

\begin{figure}[t]
    \centering
    \includegraphics[width=\textwidth]{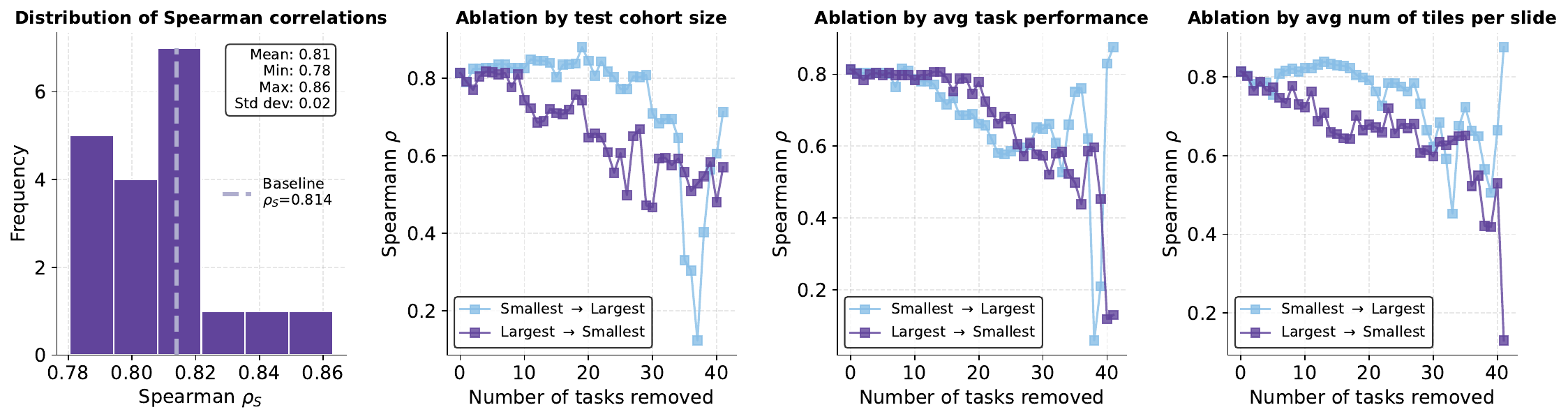}
    \caption{\textbf{Comprehensive sensitivity analysis of tile-to-slide performance correlation on ABMIL.} \textit{From left to right}: (\textit{1}) Leave-One-Model-Out  sensitivity distribution demonstrating a stable Spearman correlation ($\rho_s$) when individual foundation models are removed. The remaining plots show correlation trajectories when slide-level tasks are iteratively removed (smallest-to-largest vs. largest-to-smallest) based on: (\textit{2}) test cohort size, (\textit{3}) average task performance, and (\textit{4}) average number of tiles per slide.}
    \label{fig:sensitivity_analysis}. 
\end{figure}

\begin{figure}[!t]
\centering
  \begin{minipage}[b]{.65\linewidth}
    \centering
    \includegraphics[width=\textwidth]{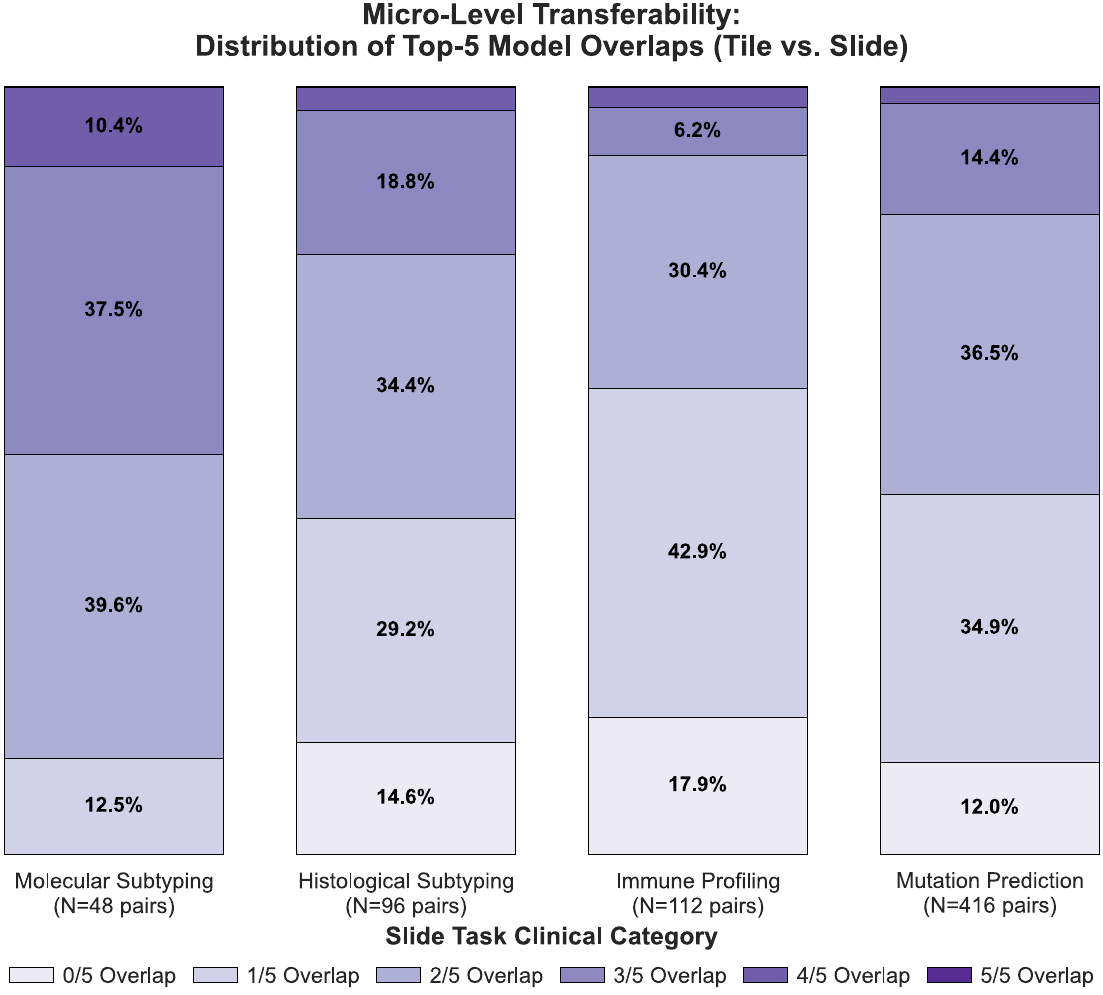}
    \captionof{figure}{\textbf{Micro-Level Transferability of Top-5 Histopathology Models.} 
    Distribution of model overlap between tile-level and slide-level tasks, stratified by slide task clinical category. For each evaluated pair of tile and slide tasks, the intersection of the top 5 highest-performing models was computed.}
    \label{fig:pairwise_tile_slide_intersection}
  \end{minipage}\hfill
  \begin{minipage}[b]{.3\linewidth}
    \centering
    {
    \scriptsize
    \captionof{table}{Rank Sum Analysis of Histopathology Models (Tile vs Mean Pooling vs ABMIL) -- T: tile-level rank, S-M: Meanpool slide-level rank, S-A: ABMIL slide-level rank}
    \begin{tabular}{>{\columncolor{gray!20}}l>{\columncolor{white}}c>{\columncolor{white}}c>{\columncolor{white}}c>{\columncolor{blue!10}}c}
    
    \toprule
    \textbf{Model} & \textbf{T} & \textbf{S-M} & \textbf{S-A} & \textbf{Sum}\\
    
    \textbf{hopt1} & 2 & 2 & 1 & \textbf{5} \\
    \textbf{uni2h} & 1 & 1 & 3 & \textbf{5} \\
    \textbf{h0mini} & 4 & 3 & 5 & \textbf{12} \\
    \textbf{keep} & 7 & 5 & 2 & \textbf{14} \\
    \textbf{virch2} & 3 & 4 & 8 & \textbf{15} \\
    \textbf{uni} & 6 & 6 & 4 & \textbf{16} \\
    \textbf{hopt0} & 5 & 7 & 7 & \textbf{19} \\
    \textbf{gigapath} & 8 & 9 & 9 & \textbf{26} \\ 
    \textbf{conch1.5} & 10 & 14 & 6 & \textbf{30} \\ 
    \textbf{conch} & 11 & 11 & 12 & \textbf{34} \\ 
    \textbf{virch} & 9 & 8 & 17 & \textbf{34} \\ 
    \textbf{hiboub} & 14 & 10 & 11 & \textbf{35} \\
    \textbf{kaiks16} & 12 & 13 & 13 & \textbf{38} \\
    \textbf{phik2} & 17 & 12 & 10 & \textbf{39} \\
    \textbf{kaikb16} & 13 & 16 & 15 & \textbf{44} \\
    \textbf{phik} & 15 & 17 & 14 & \textbf{46} \\
    \textbf{kaiks8} & 16 & 15 & 16 & \textbf{47} \\
    \textbf{quilt} & 19 & 18 & 18 & \textbf{55} \\
    \textbf{plip} & 18 & 19 & 19 & \textbf{56} \\
    
    \bottomrule
    \end{tabular}
    \label{tab:rank_sum_transposed}
    }
  \end{minipage}
\end{figure}

\myparagraph{Comprehensive sensitivity analysis of ABMIL transferability} Fig.~\ref{fig:sensitivity_analysis} shows that ABMIL tile-to-slide rank correlation is robust and not driven by any single encoder. In the leave-one-model-out analysis (left), Spearman remains near the baseline ($\rho_s \approx 0.814$) with limited variation. Task-ablation trajectories further show that correlation is more sensitive to \textbf{test cohort size} and \textbf{number of tiles per slide} than to \textbf{average slide-task performance}. Indeed, removing large-cohort tasks first, or high-tile-count tasks first, causes an earlier drop in correlation (including an approximate 10\% decrease before 10 removed tasks), whereas performance-based ablation is initially stable. This suggests transferability is supported more by statistical reliability and bag complexity than by task difficulty alone.

\myparagraph{Micro-level transferability of top-5 models} Complementing the \textit{global correlation} analysis, Fig.~\ref{fig:pairwise_tile_slide_intersection} studies the agreement in best-performing models between tile- and slide-level tasks. For each tile/slide task pair, we compute the overlap between the top-5 tile- and slide-level models. Overlap is mostly \emph{partial} (typically 1/5--3/5 shared models), indicating that tile-level benchmarks help \emph{shortlist} strong candidates even without exactly recovering the slide-level top-5.

Overlap is higher for molecular subtyping and mutation prediction and lower for immune profiling. A likely reason is that tile and slide tasks may require different encoder information: tile tasks emphasize local morphology, whereas slide tasks may depend more on tissue architecture, spatial organization, rare patterns, and long-range context. Thus, lower overlap does not contradict transferability; it highlights the \textit{limits} of using tile-level signals for exact top-model identification on heterogeneous slide tasks.

\myparagraph{Rank-sum consensus across Tile, Mean Pooling, and ABMIL}
Table~\ref{tab:rank_sum_transposed} complements the correlation analysis by summarizing model behavior via a rank-sum criterion. It reveals a stable top tier, with \texttt{hoptimus1} and \texttt{uni2h} tied for best, followed by a compact group (\texttt{h0mini}, \texttt{keep}, \texttt{virchow2}, \texttt{uni}, \texttt{hoptimus0}) that remains strong across all three settings.

Most reordering occurs in the \emph{middle} of the ranking, particularly under ABMIL. For example, \texttt{keep} and \texttt{conch1.5} improve under ABMIL, whereas \texttt{virchow} drops substantially. This confirms that tile-level benchmarking provides a strong \emph{consensus prior} for model selection, while ABMIL mainly scrambles mid-ranked models. At the lower end, \texttt{quilt} and \texttt{plip} consistently rank last, confirming that tile-to-slide transferability preserves both the top and bottom of the leaderboard.

\section{Conclusion}

In this paper, we show that tile-level benchmarking is a strong proxy for slide-level model selection in digital pathology. Across 19 foundation models, tile-level linear probing correlates strongly with slide-level performance, with higher correlation for mean pooling and slightly lower (but still clear) correlation for ABMIL, where learned aggregation introduces additional variability. Our sensitivity analyses indicate that the ABMIL correlation is stable and not driven by a single model, and that transferability is more sensitive to dataset properties (e.g., cohort size and number of tiles per slide) than to average task difficulty. The top-5 overlap analysis shows that tile-level benchmarks can help identify promising models for a new clinical slide-level task, even when the exact best models differ because tile and slide-level tasks may rely on different types of information. Overall, tile-level benchmarks provide an efficient and practical first step for pathology FM selection, while slide-level validation remains important for context-heavy tasks and for distinguishing between closely ranked models.

\myparagraph{Limitations} \textit{(i)} Our tile-level tasks are predominantly oriented toward local morphology, which may explain the lower top-5 overlap observed for immune profiling tasks that rely on broader spatial tissue architecture. \textit{(ii)} Beyond dataset scope, more expressive aggregators such as ABMIL introduce learned task-specific parameters that reshape the frozen FM's representation space, introducing aggregation-specific variability that may affect transferability for certain tasks. \textit{(iii)} Tile-level benchmarking requires laborious manual annotations. However, this remains a fixed, one-time cost (already absorbed by public benchmarks like \texttt{THUNDER}). In contrast, the recurring hardware cost of evaluating new FMs or conducting hyperparameter searches is a bottleneck that tile-level benchmarking can substantially reduce across development cycles. 

We view these remaining gaps as opportunities, and would like to emphasize that data quality might be a major factor to consider: richer, context-aware tile-level datasets will strengthen tile-to-slide transferability, further consolidating tile-level benchmarking as the first step for FM selection in digital pathology.

\myparagraph{Acknowledgments} This work has been supported by the \textit{Agence Nationale de la Recherche} through \textit{ANR-23-IAHU-0002}, \textit{ANR-21-CE45-0007}, \textit{ANR-23-CE45-0029}, \textit{ANR-23-IACL-0003} (\textit{DATAIA CLUSTER}) and the \textit{Health Data Hub} as part of the second edition of the \textit{France-Québec} call for projects \textit{Intelligence Artificielle en santé}. This work was performed using HPC resources from \textit{GENCI-IDRIS} (Grant \textit{2025-AD011015593R1}).

\myparagraph{Disclosure of Interests}
 The authors declare no competing interests.
\bibliographystyle{splncs04}
\bibliography{main}

@article{chen2024towards,
  title={Towards a general-purpose foundation model for computational pathology},
  author={Chen, Richard J and Ding, Tong and Lu, Ming Y and others},
  journal={Nature Medicine},
  year={2024},
}

@article{vorontsov2024foundation,
  title={A foundation model for clinical-grade computational pathology and rare cancers detection},
  author={Vorontsov, Eugene and Bozkurt, Alican and Casson, Adam and others},
  journal={Nature medicine},
  year={2024},
}

@article{zimmermann2024virchow2,
  title={Virchow2: Scaling self-supervised mixed magnification models in pathology},
  author={Zimmermann, Eric and Vorontsov, Eugene and Viret, Julian and others},
  journal={arXiv},
  year={2024}
}

@online{hoptimus0,
  author = {Saillard, Charlie and Jenatton, Rodolphe and Llinares-López, Felipe others},
  title = {H-Optimus-0},
  year = 2024,
  url = {https://github.com/bioptimus/releases/tree/main/models/h-optimus/v0},
}

@inproceedings{filiot2025distilling,
  title={Distilling foundation models for robust and efficient models in digital pathology},
  author={Filiot, Alexandre and Dop, Nicolas and Tchita, Oussama and others},
  booktitle={MICCAI},
  year={2025},
}

@article{lu2024visual,
  title={A visual-language foundation model for computational pathology},
  author={Lu, Ming Y and Chen, Bowen and Williamson, Drew FK and others},
  journal={Nature Medicine},
  year={2024},
}

@article{ding2024multimodal,
  title={Multimodal whole slide foundation model for pathology},
  author={Ding, Tong and Wagner, Sophia J and Song, Andrew H and others},
  journal={arXiv},
  year={2024}
}

@article{shao2025multiple,
  title={Do Multiple Instance Learning Models Transfer?},
  author={Shao, Daniel and Chen, Richard J and Song, Andrew H and others},
  journal={arXiv preprint arXiv:2506.09022},
  year={2025}
}

@article{filiot2023scaling,
  title={Scaling self-supervised learning for histopathology with masked image modeling},
  author={Filiot, Alexandre and Ghermi, Ridouane and Olivier, Antoine and others},
  journal={medRxiv},
  year={2023},
}

@article{filiot2024phikon,
  title={Phikon-v2, a large and public feature extractor for biomarker prediction},
  author={Filiot, Alexandre and Jacob, Paul and Mac Kain, Alice and others},
  journal={arXiv},
  year={2024}
}

@article{nechaev2024hibou,
  title={Hibou: A family of foundational vision transformers for pathology},
  author={Nechaev, Dmitry and Pchelnikov, Alexey and Ivanova, Ekaterina},
  journal={arXiv},
  year={2024}
}

@article{aben2024towards,
  title={Towards large-scale training of pathology foundation models},
  author={Aben, Nanne and de Jong, Edwin D and Gatopoulos, Ioannis and others},
  journal={arXiv},
  year={2024}
}

@article{karasikov2025training,
  title={Training state-of-the-art pathology foundation models with orders of magnitude less data},
  author={Karasikov, Mikhail and van Doorn, Joost and K{\"a}nzig, Nicolas and others},
  journal={arXiv},
  year={2025}
}

@article{zhou2024knowledge,
  title={A Knowledge-enhanced Pathology Vision-language Foundation Model for Cancer Diagnosis},
  author={Zhou, Xiao and Sun, Luoyi and He, Dexuan and others},
  journal={arXiv},
  year={2024}
}

@article{ikezogwo2023quilt,
  title={Quilt-1m: One million image-text pairs for histopathology},
  author={Ikezogwo, Wisdom and Seyfioglu, Saygin and Ghezloo, Fatemeh and others},
  journal={NeurIPS},
  year={2023}
}

@article{huang2023visual,
  title={A visual--language foundation model for pathology image analysis using medical twitter},
  author={Huang, Zhi and Bianchi, Federico and Yuksekgonul, Mert and others},
  journal={Nature medicine},
  year={2023},
}

@article{xiang2025vision,
  title={A vision--language foundation model for precision oncology},
  author={Xiang, Jinxi and Wang, Xiyue and Zhang, Xiaoming and others},
  journal={Nature},
  year={2025},
}

@article{oquab2023dinov2,
  title={Dinov2: Learning robust visual features without supervision},
  author={Oquab, Maxime and Darcet, Timoth{\'e}e and Moutakanni, Th{\'e}o and others},
  journal={arXiv},
  year={2023}
}

@inproceedings{radford2021learning,
  title={Learning transferable visual models from natural language supervision},
  author={Radford, Alec and Kim, Jong Wook and Hallacy, Chris and others},
  booktitle={ICML},
  year={2021},
}

@article{vaidya2025molecular,
  title={Molecular-driven Foundation Model for Oncologic Pathology},
  author={Vaidya, Anurag and Zhang, Andrew and Jaume, Guillaume and others},
  journal={arXiv},
  year={2025}
}

@article{shaikovski2024prism,
  title={Prism: A multi-modal generative foundation model for slide-level histopathology},
  author={Shaikovski, George and Casson, Adam and Severson, Kristen and others},
  journal={arXiv},
  year={2024}
}

@article{wang2024pathology,
  title={A pathology foundation model for cancer diagnosis and prognosis prediction},
  author={Wang, Xiyue and Zhao, Junhan and Marostica, Eliana and others},
  journal={Nature},
  year={2024},
}

@article{xu2024whole,
  title={A whole-slide foundation model for digital pathology from real-world data},
  author={Xu, Hanwen and Usuyama, Naoto and Bagga, Jaspreet and others},
  journal={Nature},
  year={2024},
}

@article{brancati2022bracs,
  title={Bracs: A dataset for breast carcinoma subtyping in h\&e histology images},
  author={Brancati, Nadia and Anniciello, Anna Maria and Pati, Pushpak and others},
  journal={Database},
  year={2022},
}

@article{marza2025thunder,
  title={{THUNDER}: Tile-level Histopathology image UNDERstanding benchmark},
  author={Marza, Pierre and Fillioux, Leo and Boutaj, Sofi{\`e}ne and others},
  journal={Neural Information Processing Systems (NeurIPS) D\&B Track},
  year={2025}
}

@inproceedings{gatopoulos2024eva,
  title={eva: Evaluation framework for pathology foundation models},
  author={Gatopoulos, Ioannis and K{\"a}nzig, Nicolas and Moser, Roman and others},
  booktitle={MIDL},
  year={2024}
}

@article{zhang2025accelerating,
  title={Accelerating Data Processing and Benchmarking of AI Models for Pathology},
  author={Zhang, Andrew and Jaume, Guillaume and Vaidya, Anurag and others},
  journal={arXiv},
  year={2025}
}

@article{campanella2025clinical,
  title={A clinical benchmark of public self-supervised pathology foundation models},
  author={Campanella, Gabriele and Chen, Shengjia and Singh, Manbir and others},
  journal={Nature Communications},
  year={2025},
}

@article{breen2025comprehensive,
  title={A comprehensive evaluation of histopathology foundation models for ovarian cancer subtype classification},
  author={Breen, Jack and Allen, Katie and Zucker, Kieran and others},
  journal={NPJ Precision Oncology},
  year={2025},
}

@article{neidlinger2024benchmarking,
  title={Benchmarking foundation models as feature extractors for weakly-supervised computational pathology},
  author={Neidlinger, Peter and El Nahhas, Omar SM and Muti, Hannah Sophie and others},
  journal={arXiv},
  year={2024}
}

@article{wolflein2023benchmarking,
  title={Benchmarking pathology feature extractors for whole slide image classification},
  author={W{\"o}lflein, Georg and Ferber, Dyke and Meneghetti, Asier R and others},
  journal={arXiv},
  year={2023}
}

@article{lee2025benchmarking,
  title={Benchmarking pathology foundation models: Adaptation strategies and scenarios},
  author={Lee, Jaeung and Lim, Jeewoo and Byeon, Keunho and others},
  journal={Computers in Biology and Medicine},
  year={2025},
}

@inproceedings{kang2023benchmarking,
  title={Benchmarking self-supervised learning on diverse pathology datasets},
  author={Kang, Mingu and Song, Heon and Park, Seonwook and others},
  booktitle={CVPR},
  year={2023}
}

@article{gustafsson2024evaluating,
  title={Evaluating Computational Pathology Foundation Models for Prostate Cancer Grading under Distribution Shifts},
  author={Gustafsson, Fredrik K and Rantalainen, Mattias},
  journal={arXiv},
  year={2024}
}

@article{alfasly2024foundation,
  title={Foundation models for histopathology—fanfare or flair},
  author={Alfasly, Saghir and Nejat, Peyman and Hemati, Sobhan and others},
  journal={Mayo Clinic Proceedings: Digital Health},
  year={2024},
}

@article{majzoub2025good,
  title={How Good is my Histopathology Vision-Language Foundation Model? A Holistic Benchmark},
  author={Majzoub, Roba Al and Malik, Hashmat and Naseer, Muzammal and others},
  journal={arXiv},
  year={2025}
}

@article{alfasly2025validation,
  title={Validation of histopathology foundation models through whole slide image retrieval},
  author={Alfasly, Saghir and Alabtah, Ghazal and Hemati, Sobhan and others},
  journal={Scientific Reports},
  year={2025},
}

@inproceedings{ilse2018attention,
  title={Attention-based deep multiple instance learning},
  author={Ilse, Maximilian and Tomczak, Jakub and Welling, Max},
  booktitle={ICML},
  year={2018},
}

@article{tomczak2015review,
  title={Review The Cancer Genome Atlas (TCGA): an immeasurable source of knowledge},
  author={Tomczak, Katarzyna and Czerwi{\'n}ska, Patrycja and Wiznerowicz, Maciej},
  journal={Contemporary Oncology},
  year={2015},
}

@article{edwards2015cptac,
  title={The CPTAC data portal: a resource for cancer proteomics research},
  author={Edwards, Nathan J and Oberti, Mauricio and Thangudu, Ratna R and others},
  journal={Journal of proteome research},
  year={2015},
}

@article{ehteshami2017diagnostic,
  title={Diagnostic assessment of deep learning algorithms for detection of lymph node metastases in women with breast cancer},
  author={Ehteshami Bejnordi, Babak and Veta, Mitko and Johannes van Diest, Paul and others},
  journal={Jama},
  volume={318},
  number={22},
  pages={2199--2210},
  year={2017}
}
\end{document}